# Towards the effectiveness of Deep Convolutional Neural Network based Fast Random Forest Classifier

**Mrutyunjaya Panda**
Reader, P.G.Department of Computer Science and Applications,
Utkal University, Vani Vihar, Bhubaneswar-751004, India
mrutyunjaya74@outlook.com

**Abstract**
Deep Learning is considered to be a quite young in the area of machine learning research, found its effectiveness in dealing complex yet high dimensional dataset that includes but limited to: images, text and speech etc. with multiple levels of representation and abstraction. As there are plethora of research on these datasets by various researchers , a win over them needs a lots of attention. Careful setting of Deep learning parameters are of paramount importance in order to avoid the overfitting unlike conventional methods with limited parameter settings. Deep Convolutional neural network (DCNN) with multiple layer of compositions and appropriate settings might be is an efficient machine learning method that can outperform the conventional methods in a great way. However, due to its slow adoption in learning, there are also always a chance of overfitting during feature selection process, which can be addressed by employing a regularization method called "dropout". Fast Random Forest (FRF) is a powerful ensemble classifier especially when the datasets are noisy and when the number of attributes are large in comparison to the number of instances, as is the case of Bioinformatics datasets. Several publicly available Bioinformatics dataset, Handwritten digits recognition and Image segmentation dataset are considered for evaluation of the proposed approach. The excellent performance obtained by the proposed DCNN based feature selection with FRF classifier on high dimensional datasets makes it a fast and accurate classifier in comparison the state-of-the-art.

## 1. Introduction

Today's Data Scientists working in many reputed organizations such as: Google and Microsoft etc. are paying attentions to high focus areas such as: Deep learning and Big data Analytic. It is envisaged that Deep learning is an invaluable ingredient for the big data analytic for analyzing the large volume of complex data for feature extraction at the highest abstraction levels. It is also observed that Deep learning algorithms are motivated from the artificial intelligence where the deep layered hierarchical learning is performed in the neocortex region of human brain [1]. The empirical results obtained from various researchers opines that Deep learning architecture presents some interesting classification results when applied to Computer Vision [2], Image processing [2], speech recognition [3] etc.
As the size of the datasets are growing rapidly, an efficient machine learning technique is needed that can not only address the curse of dimensionality but also generate enough simulated collisions to describe the relative likelihoods from the full feature space of the original data [4]. While Conventional Neural network is assumed to be a "shallow" classifier with a single Hidden layer unit in dealing with a complex relative likelihood function, a deep neural network with fewer hidden units may be an alternative to alleviate this problem but at the cost of slow training [5]. Afterwards, it is proposed by various researchers that using some randomized algorithms such as 'dropout" [6] , use of large training data or unsupervised stacked encoders [7] can effectively address this situation.
As pointed out, the shallow neural network can not be extended to become a multilayer neural network, Deep Convolutional neural network is proposed by LeCun [8] and are implemented with linear convolutions followed by non-linearities, over typically more than five layers. Deep Convolutional Neural Network (DCNN) which is Inspired by the mammalian visual system have recently witnesses the state-of-the-art performances for the image datasets [9] with dimensions more than 106 and for the thousands of complex classes [10].

**Problem relevance:**
The Bioinformatics datasets used in this paper are challenging for their small number of instances with relatively larger number of features and their possible interactions, which demands novel methods for performing efficient classification in terms of a fast accurate and robust classifier. To solve these issues of dimensionality with complex behavior among the features, deep learning techniques using DCNN is used for feature generation that makes ready for classification.Even though there are lots of promising classifiers are available, we consider fast Random forest(FRF), a variant of conventional random forest of Classifier for our implementation with an aim that it will deals the above issues more dynamically, which shall be evident with our experimental results discussed later.

**Goals and Objectives:**
The goal of the paper is to primarily focus on the development of suitable deep Convolutional neural network model for efficient feature selection from high dimensional and complex datasets with their suitability to applications like: Bioinformatics, handwritten digits recognition and image segmentation, at the first instant and then, we apply fast random classifier (FRF), yet again a competitive one for classification purposes. In this paper, we show that a deep learning (DCNN) scoring function can combine with one of the best learning scoring functions that is based on random forest (FRF) for best classification accuracy. We conclude with the effectiveness of our proposed approach of combining DCNN by comparing with FRF classifier with the other available researches in the relative domain and highlight the future scope of research at last.

**Scientific novelty:**
The novelty of the paper lies with:
(1) Development of an efficient feature generation technique using Deep Convolutional Neural Networks with dropouts, in complex and high dimensional datasets applied to : Bioinformatics, handwritten digit recognition and Image segmentation.
(2) We propose to use a novel fast random forest Classifier on the features extracted from the DCNN methods to obtain a fast and accurate classifier.

**2. Related Work**
Tang (2013) [11] proposed a deep architecture using support vector machine in place of Softmax activation function for classification same as that proposed by (Zhong & Ghosh, 2000 [12]; Nagi et al., 2012 [13]) with minor novel contributions in using loss from L2-SVM rather than using hinge loss.
Yuan et al. (2012) [14] used CNN architecture based on UNIPEN English character dataset and Guyon, (1994) [15] for handwritten character recognition achieved recognition rates of 93.7% and 90.2% for lowercase and uppercase characters, respectively.
Zeng et al. (2015) [16] Proposed to use Convolutional neural network on ISH images with the application of invariant feature extractors and bag-of-words method for obtaining better gene expression patterns with average AUC of $0.894 \pm 0.014$. Levi and Hassner (2015) [17] used CNN for accurate classification age and gender considering the small sample size of unconstrained images and concludes that their approach may well be find a place for large training set also.
Kanter and Veeramachaneni (2015) [18] developed deep feature synthesis, a feature extraction process that are based on human intuition with a generalized machine learning pipeline with a Gaussian copua process based approach applied to relational datasets and demand to achieve a winning score.
Raiko et al. (2015) [19] reviewed some recently used deep learning architectures and proposed a new iterative neural auto-regressive distribution estimator (NADE-k) and claim to have better prediction than the restricted Boltzmann machines and deep belief networks.
Liu et al. (2016) [20] used Deep Convolutional Neural network in the agriculture pest insects classification with proper parameter settings with dropout ratio, loss function etc. And achieved a average precision accuracy of 95.1%, better in comparison to others.
Yang et al. (2015) [21] proposed systematic feature learning method based on CNN for human activity recognition (HAR) problem and justifies that their proposed approach outperforms others related work presented in HAR algorithms with various benchmark datasets available.
Combination of Deep auto-encoders with batch mode reinforced learning algorithm is used to improve the topology of the feature space for policy planning using synthesized and real images (lange and Riedmiller, 2010) [22]
Tran, Phung and Venkatesh (2016) [23] proposed a new framework of integrating deep neural network with formal modeling of human behaviors by iteratively eliminating the least worthy items from the dataset. They apply their approach in Yahoo! Learning dataset using Gompertz distribution instead of commonly used Plackette-Luce method. Uppu et al. (2016) [24] used deep feed-forward neural network in order to effectively address the complexity of phenotype-genotype mapping those are responsible for a disease risk. They evaluated with a number of simulated datasets and conclude that they obtain significantly better predictive accuracy in comparison the other available approaches.
Abdel-Zaher and Eldeib (2015) [25] proposed to use a back-propagation model with initial weights are taken from Deep Belief Network and learning using Liebenberg Marquardt function for Wisconsin-breast cancer detection. They claim that their approach provides an average accuracy of 99.68% in comparison to the others after performing the experiments in several train-test partitions.

Cao et al. (2013) [26] suggested that due to challenges posed by electron microscopy (EM) images during image reconstruction, an automatic image segmentation approach is proposed using CNN-RF hybrid on the benchmarks for ISBI2012 EM Segmentation Challenge. They claim that their approach is effective in producing low error rates in terms of Rand Error, Warping error and Pixel error.

Emmerson and Bradley (2013) [27] pointed out the effectiveness of Deep Belief network(DBN) on medical imaging problem, conducted experiments on images obtained from Queensland Medical Laboratory (QML) AutoCyte Cervical Cancer dataset, Wisconsin Breast cancer dataset and MNIST Handwriting Recognition datasets. They compared the DBN performance with already established Naive Bayes, K-Nearest Classifier, Multilayer perceptron and Support vector machine classifiers. Finally, They found that their proposed DBN is outperformed by simple naive Bayes and K-nearest neighbor classifier with high accuracy and concluded that some good results are obtained for image datasets than the other two ones.

Extensive experiments are performed by Dittman et al. (2015) [28] on 15 imbalanced Bioinformatics datasets using Random forest classifier with random under-sampling and two post sampling methods for improving the performance of the classifier. However, they concluded that these sampling methods are not necessary in case of Random forest classifier as they won't add any more performance advancement, as the classifier is robust on its own in dealing the class imbalance problem more effectively and efficiently.

Maji et al. (2015) [29] proposed an ensemble learning based combination method using unsupervised denoising autoencoder (DAE) and DNN for detection of blood vessels in fundus color image. This is followed by supervised DNN-Random forest hybrid for the same purpose and finally, concluded to achieve an accuracy of 93.27%.

Tripathy et al. (2016) [30] proposed to use a DNN for modeling a fast and accurate classifier to to identify long non-coding RNA in cellular functions. They applied DNN into dataset collected from LNCipedia and RefSeq database and known human genome dataset and found the accuracy of 98.07% and 99% respectively. Jo et al. (2015) [31] presented a deep learning network (DN-fold) for protein fold recognition, applied to Lindahl's benchmark dataset and on a large benchmark set extracted from SCOP 1.75 consisting of about one million protein pairs, at three different levels of fold recognition. They compared the methods with 18 known ones and found their method with good accuracy.

Pratta et al. (2016) [32] proposed a CNN approach to diagnose diabetic retinopathy (DR) through color fundus image and to classify it correctly. They train the CNN with high end graphics processor unit on Kaggle dataset and claim to obtain accuracy of 75% on the validation of 5000 images.

Parameter setting also seems to be an important requirement for the success of deep learning implementation. To move in that direction, Ciresan et al. (2012) [33] stated that the classifier performance is increased when multiple neural network classifiers are used. Further, learning rate is kept at 0.45 and appropriate settings evaluated on MINST handwritten color digit dataset, in place of gray scale ones.

**Motivation:**
Despite several challenges ahead, CNN considered to be an novel, emerging area of neural network research, which seems to produce better predictive accuracy for the larger datasets.

**Contributions:**
Our contribution mainly pointed on the following:
- To uncover the suitability of application of DCNN outside Image recognition task, where it has been popularly applied by many researchers.
- We propose to use Deep Convolutional Neural Network (DCNN) as a pre-processing step to select the best feature subset.
- We use the model to be tested with the use of a fast random forest classifier.
- The effectiveness of the hybrid DCNN-FRF is tested on diversified datasets such as: Bioinformatics, Segmentation and character recognition datasets, in terms of classification accuracy, time taken to build the model.

The remainder of the paper is structured as follows: Section 3 presents an overview of the Deep learning architectures and their working principles and suitability in the machine learning application. A detailed analysis of Fast Random Forest is provided in Section 4 for effective and efficient classifications. Section 5 presents the materials and methods used for experimentation followed by experimental results and discussion in Section 6. Finally, summary of the paper is presented in Section 7 to reiterate the achievement and put some insights to carry further research in the domain of Deep learning.

## 3. Deep Learning with CNN

Even though Deep CNN is somehow similar to linear Neural Network, the main difference lies on using a "Convolution" operator as a filter that can perform some complex operation with the help of convolution kernel. It is to be noted that for smoothing an image, Gaussian kernel and for obtaining edges of an image, Canny kernel and for gradient features, Gabor kernel as filter are widely used most in image processing applications. At the same, while comparing with Autoencoder and Restricted Boltzmann Machines, it is pointed out that DCNN is intended to find a set of locally connected neurons while the others learn from a single global weight matrix between two layers.

The central idea of using DCNN is not to be serious on using predefined kernels rather to learn data specific kernels where low-level features can be translated to the high level ones and learning is done from the spatially very close neurons.

### 3.1. Training DCNN:

The training process of DCNN consists of two phases: feed forward phase and back propagation phase. In the first phase, all the tasks are passes through the input layer to the output layer and the error are computed. Based the error obtained, back propagation phase starts with bias and weight updates for minimization of the error obtained in the first phase. Several additional parameters named as hyper-parameters such as: learning rate and momentum are set properly in the range of 0 to 1. Further, number of iterations (epochs) that are required for training is also to be mentioned, for an efficient learning of the DCNN model so that the error gradient shall below the minimum acceptable threshold. The learning rate should be chosen in such a way that it should not over-fit or over-train the model. The momentum value may be chosen with trial and error method for its best adaptability to the situations. It should be noted that if we chose high learning rate and high momentum value (close to 1), then there shall be always a chance that we may skip the minima.

### 3.2. DCNN Layers and Functions:

Initially, the concepts of Deep CNN was coined by LeCun in [34]. There are three layers in Deep Convolutional neural network such as: Input layer, one or more hidden layers, fully connected layers and output layer. The pre-processing process starts at the input layer by applying the whole dataset to it. The middle layer is the hidden layer which is the heart of the DCNN layer and the number of hidden layers to be used in this stage depends largely on the input data. Convolution function is the process that involves in scanning a data filter over the dataset used. The filter is thought of as an array of 2-D data of the size 3x3,5x5 or 7x7 etc. The set of nodes obtained from the convolution process are termed as feature maps. While the dataset used is the input to the first hidden layer, the number of feature maps obtained from the previous layers becomes the input for the subsequent hidden layers. At the same time, for each layer, the number of output feature maps shall be same as the number of filter that are to be used in the convolution process. Based on the type of convolution function used in the process, the size of the output feature maps can be determined. As the feature extraction process depends on the size and number of filters used in each hidden layer, processing the large dataset with huge parameters demand a high end computing environment with large memory capabilities for providing optimum performance. In this, use of random filter is a good choice that are represented in the range of [-1 to +1]. The output from convolution process is applied to an activation function for linear or non-linear transformation across different layers so that the output falls in the range of [-1,+1] or [0,1]. Even though several activation functions available, we use rectified linear unit activation function for our analysis which says that for input x and output y, $f(y) = 1$, if $x > 0$, and $f(y) = 0$, otherwise. Finally, pooling function is used (here, we use max-pooling), for down sample and down size the input features map to half of its original size. Dimensionality reduction on the datasets shall be achieved by end of this stage.

Fully connected layer is the last but one to the DCNN layers where the result is obtained as a single vector after consolidating the previous layers single-node output feature maps. As this stage leads to perform classification on the dataset, proper weights and bias are applied in this layer so that the cost of mis-classification falls below a certain threshold or else the back-propagation process to get initiated for better weight and bias updates across layers for error minimization. Finally, the output layer presents the desired output. Different hyper-parameters such as learning rate and momentum are then specified to have a faster convergence.

In order to suitably address the slow training of large neural network while dealing with overfitting problems, some randomly chosen units are dropped (popularly known as Dropout) during the DCNN training [58].

## 4. Fast Random Forest (FRF)

After pre-processing is done with large datasets using DCNN process, the goal is to develop some efficient machine learning algorithm that can help to speed up the classification process and address the memory constraints effectively. We propose to use Fast random Forest (Wright and Ziegler; 2016; )[35] in our analysis looking into its strength in dealing with complex high dimensional datasets where the number of attributes are larger than the number of instances.(Boulesteix et al.; 2015) [36]

### 4.1. Algorithm description

Random Forest algorithm is a fast, highly flexible and fully non-parametric method which is an ensemble of decision tress that are built from a subset of data (Breiman; 2001) [37].The other data instances those are not considered for decision tree construction are termed as out-of-bag(OOB) instances. The Gini index is used as splitting criteria for attribute selection.The proposed fast random forest algorithm uses one-time sorting by separating the lists for each predictor output in order to reduce the computational burden.Here, sorted list are placed in main memory one at a time so that a class list with at least one list of sorted indices must always fit into main memory. The main motivation in using this FRF algorithm for classification of large, complex datasets in this paper, is mainly because of its ability to rank the attributes according to their predictive importance, which suits to our applications in Bioinformatics datasets, character recognition and segmentation datasets.

## 5. Materials and Methods

This section will perform three group of experiments considering real-world datasets collected from three different application domains. The proposed methodology is shown in Figure 1.

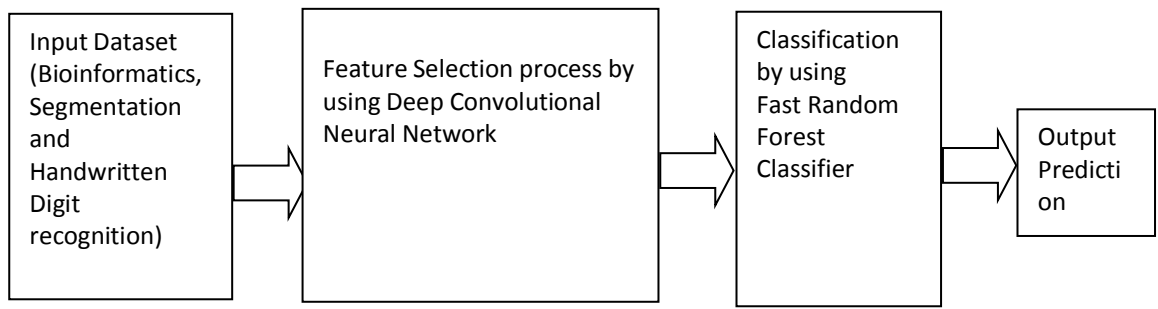

**Figure 1: Proposed methodology**

This section describes three different experiments on real world datasets in the application areas of Bioinformatics, handwritten digit recognition and Image segmentation. Further, the performance comparisons are conducted with other available approaches in the literature in terms of accuracy.

### 5.1. Data Set

We use Bioinformatics datasets for understanding the effectiveness of our proposed classification . We use Arrhythmia, Leukemia, Lymphoma and prostrate cancer datasets for the experimentation. These are interesting as the datasets consists of more number of attributes than the number of instances. The number of attributes and instances for Arrhythmia, Leukemia, Lymphoma and prostrate cancer datasets are 280,452; 7130,38; 4027,96 and 12601,102 respectively.

The second application datasets relate to Handwritten digit recognition dataset that include: Opt-digit, pen-digit and KDD Cup Japanese vowel datasets. Here, the number of attributes and samples are: 65, 5620; 17,10992; and 15,4274 respectively.

Third application is in image segmentation domain, where Segment and Landsat datasets are used for the analysis of the proposed approach. The segmentation data consists of 10 attributes and 2310 instances; the satellite image dataset (Landsat) consisting of 37attributes and 4435 instances.

To evaluate the performance of different methods, we use feature selection process by using Deep convolution neural network. We use hidden two Convolutional layers in terms of (Number of feature maps, patch width, patch height, pool-width and pool-height) as: 20-5-5-2-2 and 100-5-5-2-2 and then with 6-3-3-2-2 and 12-3-3-2-2 for the performance comparison on diverse application domains. We use DCNN with batch size=100, learning rate=0.95, hidden layers drop out rate=0.5, input layer dropout rate =0.2, number of epochs=100 to train on the entire dataset.

For fast random forest, we chose to calculate the out-of-bag (OOB) error with number of batches for classification =100.

## 6. Experiment Setup and Results

All the experiments are conduced in a Intel core i5 processor with 1TB HDD, 2GB RAM, PC using Windows 8.1 under Java environment. We use 5-fold cross validation for testing the approach. The results obtained are shown in Table 1, Table 2 and Table 3 for DCNN applications to diverse datasets. Table 4, Table 5 and Table 6 presents the results obtained from the combination of DCNN and fast random forest classifiers, with an intention to further improve the accuracy obtained in the case of single DCNN classifier. We then compare the results with others in Table 7, Table 8 and table 9. Finally, statistical significance test were carried out to conclude the most significant classifiers in all the application domains.

### 6.1. Experimental Results-1: CNN as classifier with 20-5-5-2-2 and 100-5-5-2-2

**Table 1: DCNN with Bioinformatics dataset**

| Sl.No. | Bioinformatics Dataset | No. Of Attributes | No. Of Instances | Time taken to build the model in seconds | Accuracy in % |
|---|---|---|---|---|---|
| 1 | Lymphoma | 4027 | 96 | 78.56 | 29.17 |
| 2 | Arrhythmia | 280 | 452 | 305.09 | 65.58 |
| 3 | Leukemia | 7130 | 38 | 4.49 | 100 |
| 4 | Dermatology | 35 | 366 | 4.51 | 30.61 |

**Table 2: DCNN with Image dataset**

| Sl.No. | Image dataset | No. Of Attributes | No. Of Instances | Time taken to build the model in seconds | Accuracy in % |
|---|---|---|---|---|---|
| 1 | Segmentation | 10 | 2310 | 0.73 | 89.34 |
| 2 | Landsat | 37 | 4435 | 4.64 | 23.05 |

**Table 3: DCNN with Handwritten digit dataset**

| Sl.No. | Handwritten Digit Recognition dataset | No. Of Attributes | No. Of Instances | Time taken to build the model in seconds | Accuracy in % |
|---|---|---|---|---|---|
| 1 | Japanese | 15 | 4274 | 162.43 | 13.69 |
| 2 | Opt digit | 65 | 5620 | 173.25 | 9.57 |
| 3 | Pen digits | 17 | 10992 | 35.68 | 9.98 |

### 6.2. Experimental Results-2: CNN with 6-3-3-2-2 and 12-3-3-2-2+FRF Classifier

**Table 4: DCNN+FRF with Bioinformatics dataset**

| Sl.No. | Bioinformatics Dataset | No. Of Attributes | No. Of Instances | Random Features | OOB error | Time taken to build the model in seconds | Accuracy in % |
|---|---|---|---|---|---|---|---|
| 1 | Lymphoma | 4027 | 96 | 12 | 0.1979 | 0.86 | 81.25 |
| 2 | Arrhythmia | 280 | 452 | 9 | 0.3142 | 0.95 | 70.78 |
| 3 | Leukemia | 7130 | 38 | 13 | 0.1351 | 0.34 | 81.58 |
| 4 | Dermatology | 35 | 366 | 6 | 0.0273 | 0.05 | 95.91 |

**Table 5: DCNN+FRF with Image dataset**

| Sl.No. | Image dataset | No. Of Attributes | No. Of Instances | Random Features | OOB error | Time taken to build the model in seconds | Accuracy in % |
|---|---|---|---|---|---|---|---|
| 1 | Segmentation | 10 | 2310 | 5 | 0.0186 | 0.73 | 97.49 |
| 2 | Landsat | 37 | 4435 | 4 | 0.021 | 2.09 | 91.28 |

**Table 6: DCNN+FRF with Handwritten digit dataset**

| Sl.No. | Handwritten Digit Recognition dataset | No. Of Attributes | No. Of Instances | Random Features | OOB error | Time taken to build the model in seconds | Accuracy in % |
|---|---|---|---|---|---|---|---|
| 1 | Japanese | 15 | 4274 | 4 | 0.0236 | 1.88 | 97.31 |
| 2 | Opt digit | 65 | 5620 | 7 | 0.019 | 2.5 | 98.21 |
| 3 | Pen digits | 17 | 10992 | 5 | 0.0089 | 4.19 | 99.06 |

While comparing with the results obtained from our approach with different Deep Convolutional neural Network settings (Table 1, Table 2 and Table 3 with Table 4, Table 5 and Table 6) in Experimental results 1 and Experimental results 2, one can argue that choosing the right setting is most important to see that DCNN with FRF produce accurate classifier for all datasets. The result confirms that DCNN with 6-3-3-2-2 and 12-3-3-2-2+FRF enhances the accuracy for all the datasets except leukemia data in comparison to DCNN as single classifier with 20-5-5-2-2 and 100-5-5-2-2 without pre-processing. Further, extensive search is made to find out the relevant papers those using deep learning for all the application domain dataset for comparison with our obtained result to provide an amicable solutions. The comparisons are provided in Table 7, Table 8 and Table 9 for Bioinformatics, Image and Handwritten digit recognition dataset respectively.

**Table 7: Comparison with state of the art for Bioinformatics dataset**

| A. Arrhythmia | |
|---|---|
| Algorithm | Accuracy (%) |
| Autoencoder [45] | 52.6 |
| SVDD [45] | 55.7 |
| ELM [45] | 52.8 |
| DCNN+FRF [ours] | **70.78** |
| **B. Lymphoma** | |
| Algorithm | Accuracy (%) |
| Hybrid BPNN+GA [39] | **85.65** |
| DCNN+FRF [ours] | 81.25 |
| **C. Leukemia** | |
| Algorithm | Accuracy (%) |
| aRBm [38] | **96.1** |
| DBN [40] | 98.6 |
| Hybrid BPNN+GA [39] | 89.33 |
| DCNN+FRF [ours] | 81.58 |
| **D. Dermatology** | |
| Algorithm | Accuracy (%) |
| AECOC [51] | **96.99** |
| DCNN+FRF [ours] | 96.2 |

**Table 8: Comparison with others for Image dataset**

| A. Segmentation Data | |
|---|---|
| Algorithm | Accuracy (%) |
| AECOC [51] | 93.29 |
| US-ELM [52] | 74.5 |
| DCNN+FRF (ours) | **97.49** |
| **B. Landsat- Satellite Image Dataset** | |
| Algorithm | Accuracy (%) |
| MLP-BPNN [54] | 89.3 |
| AECOC [51] | 85.72 |
| ANOVA-KERNEL SVM [57] | 82.87 |
| DCNN+FRF (ours) | **91.28** |

**Table 9: Comparison with others for Handwritten Digit recognition datasets**

| A. OptDigit Dataset | |
|---|---|
| Algorithm | Accuracy (%) |
| LDNN [55] | 97.71 |
| AECOC [51] | 97.95 |
| ABC-Boost [50] | 97.5 |
| AdaBoost.MH [56] | 98 |
| DCNN+FRF (ours) | **98.21** |
| **B. PenDigit** | |
| Algorithm | Accuracy (%) |
| LDNN [55] | 98.2 |
| STKG-PSVM-k5-M [41] | 98.94 |
| Gaussian Interpolation Kernel [42] | 97.3 |
| Discrete SVM [43] | 97.2 |
| ABC-Boost [50] | 72 |
| AdaBoost.MH[56] | 97.9 |
| AECOC [51] | 96.84 |
| DCNN+FRF[ours] | **99.06** |
| **C. KDD Cup Japaneses Vowels** | |
| Algorithm | Accuracy |
| STKG-PSVM-k5-M [41] | **98.75** |
| Gaussian Interpolation Kernel [42] | 97.29 |
| Discrete SVM [43] | 96.7 |
| DCNN+FRF (ours) | 97.31 |

From all these analysis, it is quite evident that our DCNN+FRF model is either the most accurate one or very close to the most accurate ones. It can also be seen that the proposed DCNN+FRF performs well when the number of instances in the dataset are large. Hence, it may well be suited to perform better when can effectively be applied to big data analytic.

### 6.3. Statistical Significance Test:
Finally, we perform two tailed pairwise t-test as a part of statistical significance test for measuring the most efficient classifiers for all the dataset used in the paper. It is customary to mention here that non-parametric two tailed paired t-tests compare two sample means (here accuracy) in order to find the difference between them. One considered to be statistically significant then the other if the p-value falls below 0.05 at 95% significance level.

It is observed from the statistical test that all are equally significant and hence cannot reject the null hypothesis.

## 7. Conclusion and Future Scope

In this paper, we have shown the effectiveness of deep Convolutional neural network as a feature selection approach with the use of fast random classifier to perform an efficient classification task. We have demonstrate the effectiveness of our approach by performing extensive simulation over three most promising application domains such as: Bioinformatics, Image and Handwritten digit recognition. The comparison with the others related work with test of statistical significance demands the suitability of our proposed classifier in a promising way. In future, we shall concentrate our work to improve the classification with more developments in DCNN architecture with application to some big complex yet noisy datasets.